\pdfoutput=1
\documentclass[fleqn,10pt]{wlscirep}
\usepackage[utf8]{inputenc}
\usepackage[T1]{fontenc}
\usepackage{graphicx}
\usepackage{booktabs}
\usepackage{array}
\usepackage{xurl}

\title{Parameter-Efficient Conciseness Adaptation: Differentiable Length
Control for Concise Clinical Reasoning in Smaller Language Models}

\author[1]{Tanzeel Abbas}
\author[1,*]{Ahmad Kamran Malik}
\author[2]{Nafees Qamar}
\author[1]{Sana Liaqat}

\affil[1]{Department of Computer Science, COMSATS University Islamabad, Islamabad, Pakistan}
\affil[2]{School of Health and Behavioral Sciences, Bryant University, Smithfield, Rhode Island 02917, United States}

\affil[*]{ahmad.kamran@comsats.edu.pk (corresponding author)}


\begin{abstract}
Smaller language models (SLMs) make it feasible to deploy clinical natural
language processing where large proprietary models are too costly, and
parameter-efficient fine-tuning allows them to be specialised on commodity
hardware. One obstacle persists after training: on medical tasks
these models answer at far greater length than the clinical question requires. In
clinical decision support and point-of-care use, where time is short, such
verbosity is hard to act on even when the underlying reasoning is sound. Existing
length-control methods depend on preference optimisation, reward models, or
proprietary teachers that undercut the accessibility making SLMs attractive. We introduce Parameter-Efficient Conciseness Adaptation (PECA),
which adds a single differentiable conciseness term to the QLoRA supervised
fine-tuning objective. Because the token count is discrete and non-differentiable,
PECA instead raises the end-of-sequence probability once generation exceeds a
token budget, a smooth signal optimised during ordinary training, with no preference data, no
auxiliary model, and a single commodity GPU. On a
1.7-billion-parameter model fine-tuned on the medical-o1-reasoning-SFT corpus,
PECA reduces mean response length by 72.8\% while improving multiple-choice
diagnostic accuracy from 40.8\% to 44.0\%. A sweep over the conciseness strength
traces an interpretable accuracy-to-length frontier for balancing brevity and
diagnostic completeness.
\end{abstract}

\begin{document}

\flushbottom
\maketitle
\thispagestyle{plain}
\pagestyle{plain}

\section*{Introduction}\label{sec1}

Healthcare is absorbing language technology at a remarkable pace.
Transformer-based large language models (LLMs) already assist with clinical
question answering, medical documentation, patient triage, and decision support,
and on a number of diagnostic benchmarks their scores now rival, or even surpass,
those of trained clinicians. This ability is largely inherited from large-scale
pretraining, which yields representations general enough to carry over from
ordinary text into the more specialised vocabulary of medicine [1, 2]. In the
clinic this strength cuts both ways. The models that score best are usually the
ones a hospital can least easily run, audit, or host on its own servers, and the
reason is not only their price. Patient data cannot leave the building, and the
largest systems are precisely the ones that are hardest to keep inside it.

Hardware is the first wall. An LLM with tens of billions of parameters may need
hundreds of gigabytes of memory and dedicated accelerators to serve, and that
bill by itself is enough to place it out of reach for most hospitals, clinics,
and the smaller research groups around them [3]. It is this practical ceiling,
more than any lack of capability, that has drawn attention toward smaller
language models (SLMs) of roughly one to thirteen billion parameters. Such models
retain a good deal of what makes their larger counterparts useful, yet they
remain light enough to run on commodity hardware and, at least in principle, on
machines that never leave the provider's own premises [12]. For clinicians,
low-resource hospitals, and medical research groups working under tight budgets
and strict data-governance constraints, SLMs are a realistic way to deploy
advanced clinical language technology, and parameter-efficient fine-tuning (PEFT)
has been central to making that possible.

Adapting an SLM to a specialised domain exposes a problem that accuracy
benchmarks tend to hide. PEFT solves the cost of training; what it leaves
untouched is a habit that surfaces afterwards, namely that domain-adapted SLMs
answer at much greater length than the question warrants. The issue is sharpest
wherever information has to be delivered fast and without ambiguity. Clinical NLP
has made this very clear. Compact models fine-tuned on medical reasoning have
matched or beaten specialist physicians on diagnostic benchmarks [18], yet those
same clinicians note that the verbosity of the explanations makes them
impractical at the bedside, where a decision cannot wait for a long exposition.
Being correct is necessary, but an answer that takes longer than the time
available has failed regardless of how accurate it is.

This verbosity is a gap the PEFT literature has mostly passed over. LoRA [2] and
its quantised form QLoRA [12] save a great deal by confining weight updates to
low-rank matrices, but neither shapes how long the outputs run. Length follows
from the training distribution alone, never from an explicit objective. The
methods that do target length usually sit at the preference-optimisation stage
[14, 32, 33], which needs paired preferences or a learned reward model, or they
distil from a stronger proprietary teacher [11], which needs API access. Either
way the dependencies cut against the low-resource, open case for using SLMs at
all.

We close this gap with Parameter-Efficient Conciseness Adaptation (PECA). It
folds conciseness straight into the supervised fine-tuning objective through one
auxiliary term, with no preference data, no auxiliary model, and nothing external
at inference. The design hinges on how length is penalised. A penalty written
over the number of generated tokens cannot be optimised by gradient descent,
since token count is discrete and its gradient with respect to the parameters is
zero almost everywhere. So PECA acts on the model's probability of stopping
instead: once generation passes a token budget, the objective raises the
probability assigned to the end-of-sequence token, a continuous signal that
shapes stopping behaviour during normal training. The procedure adds almost
nothing to the cost of QLoRA fine-tuning and reproduces on a single commodity
GPU.

\noindent The main contributions of this paper are as follows.
\begin{itemize}
    \item We identify post-training verbosity as a practical limitation of
    PEFT-adapted medical SLMs and treat it as a problem separate from task
    accuracy.
    \item We propose PECA, a preference-free and fully differentiable
    conciseness objective for QLoRA supervised fine-tuning. We show why a naive
    penalty on the discrete token count cannot work, and we give a correct
    end-of-sequence surrogate that is easy to implement and runs on consumer
    GPUs.
    \item We provide a systematic evaluation across four values of the
    conciseness strength $\lambda$, revealing a clear Pareto frontier between
    response length and diagnostic accuracy, and identifying $\lambda=0.25$ as
    the setting that achieves 72.8\% length reduction with improved MedMCQA
    accuracy.
\end{itemize}

\section*{Related Work}\label{sec2}

Full fine-tuning is expensive, and a large body of work has grown up around
avoiding it. The account below follows that work from the early selective-update
proposals through to recent hybrid designs, then turns to a question it has
largely neglected: how long the model's outputs actually are.

\subsection*{Foundations of Parameter-Efficient Fine-Tuning}\label{subsec2a}

Early work on parameter efficiency focused on restricting which weights are
updated during fine-tuning. Ben-Zaken et al. [3] introduced BitFit, which
modifies only the bias terms, touching roughly 0.1\% of all parameters. Effective
on classification, it reached only 61\% of full fine-tuning performance on
machine translation, revealing the limits of very sparse updates for generative
tasks.

Houlsby et al. [2, 13] addressed this by inserting small trainable modules between
the frozen transformer layers. Touching just 3 to 4\% of the parameters, these
adapters still recover 85 to 90\% of full fine-tuning performance on multitask
benchmarks. The catch is at inference. Every adapter sits in the forward path and
has to be run in turn, which drives latency up by 23 to 40\%, far too much when an
answer is needed on the spot.

Low-Rank Adaptation (LoRA), from Hu et al.\ [2], is what eventually stuck. Rather
than update a weight matrix directly, LoRA learns two small matrices whose product
approximates the change. That leaves only 0.1 to 0.6\% of the parameters
trainable, and yet on GPT-2, GPT-3, RoBERTa, and DeBERTa the method matches full
fine-tuning and occasionally edges past it. Its real advantage is that the two
matrices merge back into the base weights once training ends, so inference pays no
extra cost at all [2, 17]. The idea has since been pulled in several directions.
RaLo [25] varies the adapter rank as training proceeds; HydraLoRA [28] makes the
two sides asymmetric to squeeze out more efficiency; and the recipe has escaped
NLP entirely, driving language-specific translation in low-resource settings [5]
and even turning up in image-restoration work [27].

Prefix tuning [7, 8] takes yet another route, prepending trainable task vectors to
the input and adjusting 0.1 to 0.3\% of parameters while holding on to about 87\%
of full fine-tuning efficacy on models such as GPT-2. Its weakness in multi-task
use is a sharp sensitivity to prefix length [9].

\subsection*{Advanced Integration and Hybrid Approaches}\label{subsec2b}

Dettmers et al. [12] built QLoRA by extending LoRA with 4-bit quantisation,
cutting memory use by more than 94\% and so allowing 65-billion-parameter models
to be fine-tuned on a single 48~GB GPU with no loss of performance [12]. PEFT
greatly increases Phi-1.5 when answering scientific questions: BLEU scores
increase from 0.0027 to 0.132, ROUGE-L from 0.152 to 0.235, and BERT F1 from
0.5266 to 0.6002. In medical and legal QA, He et al.'s hybrids [3] that combine
dynamic rank allocation and chain-of-thought retrieval-augmented generation
improve BLEU scores by up to 37.5\% over standard LoRA. Quantisation has pushed
this further still, letting biomedical summarisation run on low-power devices
where memory would otherwise be a hard constraint [26].

A different idea comes from Agarwal et al.\ [23], who unmask parameters a few at a
time as PEFT proceeds, buying efficiency and a measure of interpretability at
once. QLoRA, for its part, earns its keep chiefly through memory savings that
arrive without a quality penalty [8, 12]. Nor does everything hinge on
architecture: well-chosen prompts, layered on top of PEFT, have been shown to
raise task performance for free [10].

\subsection*{Length Control and Conciseness in Language Generation}\label{subsec2c}

A quite separate literature concerns itself with how long a model's output is
rather than whether it is correct. The bulk of this work operates at the
preference-optimisation stage. Direct Preference Optimization and its relatives
have a well-documented tendency to reward longer answers, and several methods try
to correct for it: Park et al.\ [33] pull length apart from quality and bolt a
length regulariser onto the DPO objective, SimPO [32] instead adopts a
length-normalised, reference-free reward, and Length-Controlled Margin-based
Preference Optimization (LMPO) [19] adds a length-aware margin for the same
purpose. Reinforcement learning offers a parallel and much newer route to the
problem. SmartThinker [37] swaps the fixed length rewards used by earlier
GRPO-style methods for a calibration that adapts to problem difficulty, precisely
so as to avoid the over-compression a rigid penalty tends to cause, something we
too run into once the conciseness strength is pushed high. All of these ideas
work, but each one leans on preference pairs, a reward model, a reference policy,
or RL rollouts, and every one of those requirements sits awkwardly with the
low-resource, preference-free scenario we are after. Closest to our objective,
Belligoli et al. [31] control summary length by giving more weight to the
end-of-sequence (EOS) token inside the cross-entropy loss, a fully differentiable
mechanism that works with supervised data alone. The present work extends this
principle into the QLoRA parameter-efficient fine-tuning setting, applying a
budget-gated EOS term to clinical chain-of-thought models where preference data
and external teachers are unavailable.

\subsection*{Domain-Specific Applications}\label{subsec2d}

Clinical and biomedical NLP has adopted PEFT extensively. Specialised adaptations
such as Med-PaLM and BioBERT draw on QLoRA, prefix tuning, and reinforcement
learning from human feedback to improve diagnostic, educational, and documentation
tasks [6]. Liu et al. [29] apply parameter-efficient instruction tuning to
biomedical named entity recognition, obtaining strong extraction accuracy at low
computational overhead. Domain resources such as consumer medical question
answering datasets have further supported this line of work [16]. Li et al. [24] show that small-parameter LLMs can be
fine-tuned effectively for bilingual biomedical tasks across languages, and more
recent work has carried PEFT into the multimodal clinical setting, as in PeFoMed
[36] for medical visual question answering and radiology report generation, and
other recent systems fine-tune general language models directly for grounded
clinical diagnosis [38]. The
medical chain-of-thought corpus used in the present study originates from the
HuatuoGPT-o1 effort [34], which builds strong medical reasoning ability through
verifier-validated reasoning traces; that line of work, however, optimises
reasoning quality and does not address the verbosity of the resulting outputs,
which is precisely the gap PECA targets.

\subsection*{Limitations in Existing PEFT Approaches}\label{subsec2e}

Despite these advances, PEFT methods remain susceptible to domain-specific
inconsistencies [1, 4] and to catastrophic forgetting in multi-task settings,
where shared adapters can degrade performance by as much as 14\% [3, 4]. A more
fundamental omission, however, concerns output quality rather than accuracy.
Verbosity in particular has received little systematic attention: a fine-tuned
SLM may reason competently and still generate responses too long to be acted on
under time pressure [18]. PEFT has also been studied as an enabler of foundation
models in federated learning [30], which underscores its breadth across
deployment settings. The practical value of PEFT in high-stakes domains is
constrained as much by this post-training communication failure as by any
shortfall in raw accuracy.

\subsection*{Positioning of This Study Within Prior Work}\label{subsec2f}

Every one of these methods has something to recommend it. LoRA balances quality
against cost, QLoRA earns its place when resources are scarce, and adapters make
multitasking straightforward. None, though, ties parameter-efficient adaptation
to post-training conciseness in one procedure, and that gap is widest for the
sub-2B SLMs we care about. Length-control work stops at the preference-optimisation
stage and assumes preference data or a reward model is on hand; PEFT work chases
accuracy and looks no further. Our contribution bridges the two, attaching a
differentiable EOS-budget conciseness term to QLoRA fine-tuning and charting a
length-versus-accuracy Pareto frontier that a clinical deployment can tune to
taste.

\section*{Results}\label{sec:results}

Two questions drive the evaluation. Q1 asks whether PECA shortens outputs by an
amount that actually matters next to plain QLoRA fine-tuning. Q2 asks how
conciseness strength plays against diagnostic accuracy, and where along that curve
a practitioner ought to sit.

\subsection*{Experimental Setup}\label{subsec:setup}

Every condition runs on the same Qwen3-1.7B-Instruct backbone with the QLoRA
settings of Table~\ref{tab:lora-config}, decodes greedily up to a 128-token cap,
and is scored on one shared held-out split of 1,000 examples. Only the training
objective and $\lambda$ vary. We evaluate response length on the full 1,000 test
items and diagnostic accuracy on 1,000 MedMCQA validation items.

\subsection*{Baseline and PECA Conditions}\label{subsec:conditions}

We compare four conditions:
\begin{enumerate}
    \item \textbf{Vanilla QLoRA SFT}: standard cross-entropy fine-tuning
    (Eq.~\ref{eq:ce}), the primary control.
    \item \textbf{PECA ($\lambda=0.25$)}: light conciseness pressure.
    \item \textbf{PECA ($\lambda=0.50$)}: moderate conciseness pressure.
    \item \textbf{PECA ($\lambda=1.00$)}: strong conciseness pressure.
\end{enumerate}

\subsection*{Evaluation Metrics}\label{subsec:metrics}

\textbf{Conciseness (Q1).} Mean and median response length in tokens;
percentage length reduction relative to vanilla SFT; response length
distribution (Fig.~\ref{fig:length_dist}).

\noindent\textbf{Quality (Q2).} Reference-based similarity to gold responses
(ROUGE-L, BERTScore-F1); exact-match accuracy on 1,000 MedMCQA
multiple-choice questions, which cannot be inflated by writing longer answers
and is therefore the most reliable measure of preserved diagnostic capability.
BERTScore [20], which measures semantic similarity through contextual embeddings
rather than surface overlap, is reported alongside ROUGE-L for this reason.

\subsection*{Main Results}\label{subsec:main}

Table~\ref{tab:main} reports the full comparison, which is consistent across the
$\lambda$ sweep.

\begin{table}[htbp]
\centering
\caption{Main results across all conditions. Length reduction is computed
relative to vanilla SFT. MCQ accuracy is exact-match on 1,000 MedMCQA items.}
\label{tab:main}
\begin{tabular}{lccccc}
\toprule
Condition & Mean len & Len red. & ROUGE-L & BERTScore & MCQ acc. \\
\midrule
Vanilla SFT           & 256.0 & 0.0\%  & 0.162 & 0.129 & 40.8\% \\
PECA ($\lambda=0.25$) & 69.7  & 72.8\% & 0.155 & 0.118 & \textbf{44.0\%} \\
PECA ($\lambda=0.50$) & 66.2  & 74.1\% & 0.154 & 0.117 & 31.9\% \\
PECA ($\lambda=1.00$) & 65.2  & 74.5\% & 0.152 & 0.113 & 4.6\%  \\
\bottomrule
\end{tabular}
\end{table}

Two regularities in Table~\ref{tab:main} deserve separate treatment, since they
behave quite differently as $\lambda$ increases. Length is the well-behaved one.
Adding the conciseness term compresses the average response to roughly a quarter
of its original size, and it does so almost regardless of how hard we push: the
reduction is already 72.8\% at $\lambda=0.25$ and creeps up only to 74.1\% and
74.5\% thereafter. ROUGE-L and BERTScore, for their part, hardly budge from
baseline, which tells us the cuts fall on padding rather than on anything
load-bearing.

Accuracy behaved differently. At $\lambda=0.25$, MedMCQA accuracy was 44.0\%,
compared with 40.8\% for vanilla SFT, a gain of 3.2 points, so the shorter answers
were more accurate rather than less. At $\lambda=0.50$ accuracy fell to 31.9\%, and
at $\lambda=1.00$ it fell to 4.6\%; at the highest setting most responses were only
one or two tokens long. Accuracy therefore rose at low $\lambda$ and dropped sharply
at high $\lambda$, while length changed little across the three settings.

\subsection*{Conciseness versus Accuracy Trade-off}\label{subsec:pareto}

\begin{figure}[ht]
\centering
\IfFileExists{pareto.png}{\includegraphics[width=0.72\linewidth]{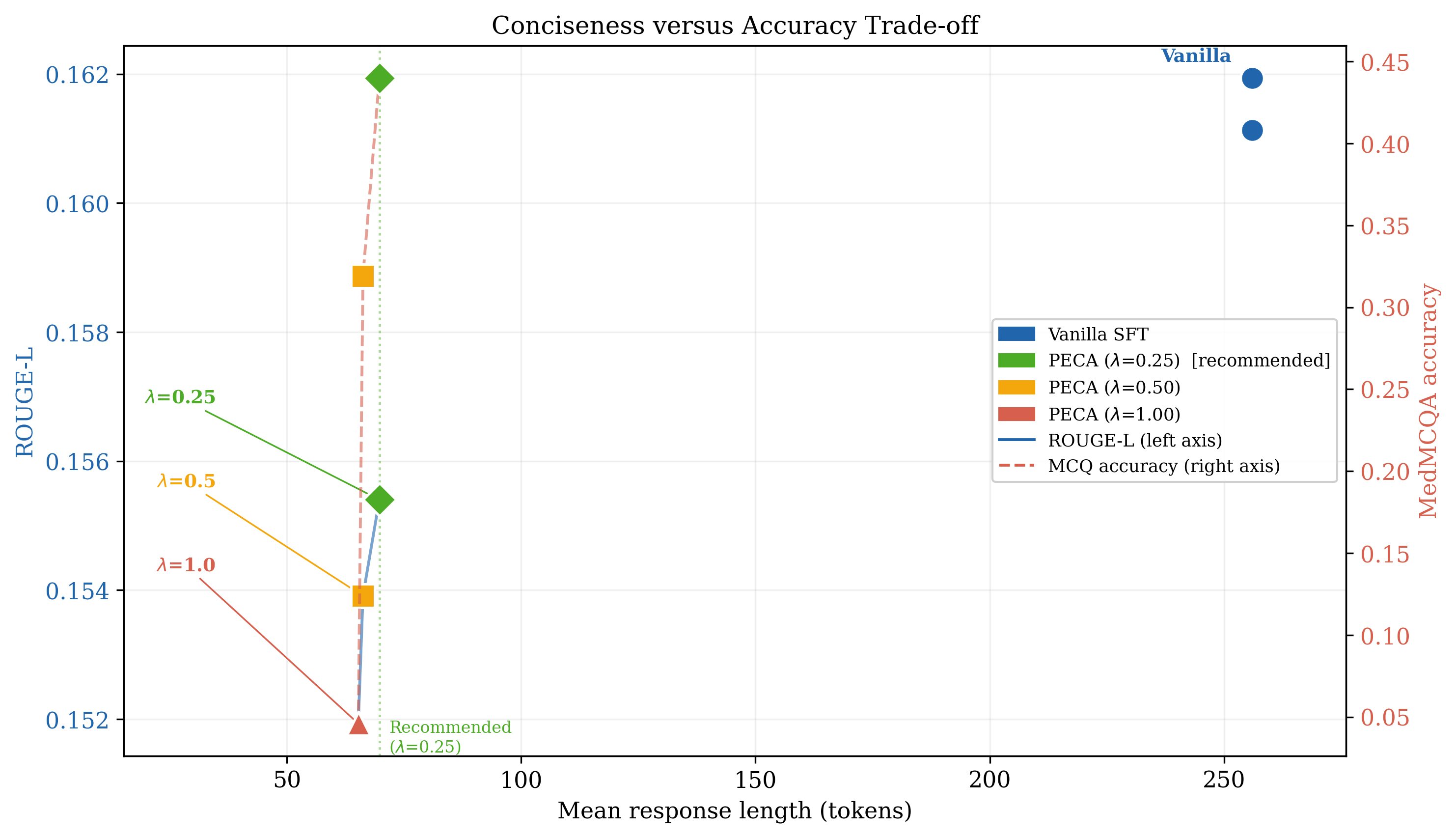}}{\fbox{\parbox[c][4cm][c]{0.8\linewidth}{\centering \textit{Figure file \texttt{pareto.png} not found.}\\[4pt]\small Upload this file to the project to render the figure.}}}
\caption{The conciseness--accuracy trade-off as $\lambda$ varies. At
$\lambda=0.25$ the model wins on both length and accuracy against vanilla SFT,
landing on the knee of the Pareto frontier.}
\label{fig:pareto}
\end{figure}

We plotted accuracy against mean length for the four conditions
(Fig.~\ref{fig:pareto}). Vanilla SFT used 256 tokens and reached 40.8\% accuracy.
At $\lambda=0.25$ the mean length was about 70 tokens and accuracy was 44.0\%, its
highest value. Raising $\lambda$ to 0.50 shortened the mean length only to 66 tokens
but lowered accuracy to 31.9\%, and $\lambda=1.00$ lowered it further to 4.6\%. Only
$\lambda=0.25$ improved on vanilla SFT in both length and accuracy at the same time,
so we use it in the rest of the paper.

\subsection*{Response Length Distribution}\label{subsec:lengthdist}

\begin{figure}[ht]
\centering
\IfFileExists{length_distribution.png}{\includegraphics[width=0.72\linewidth]{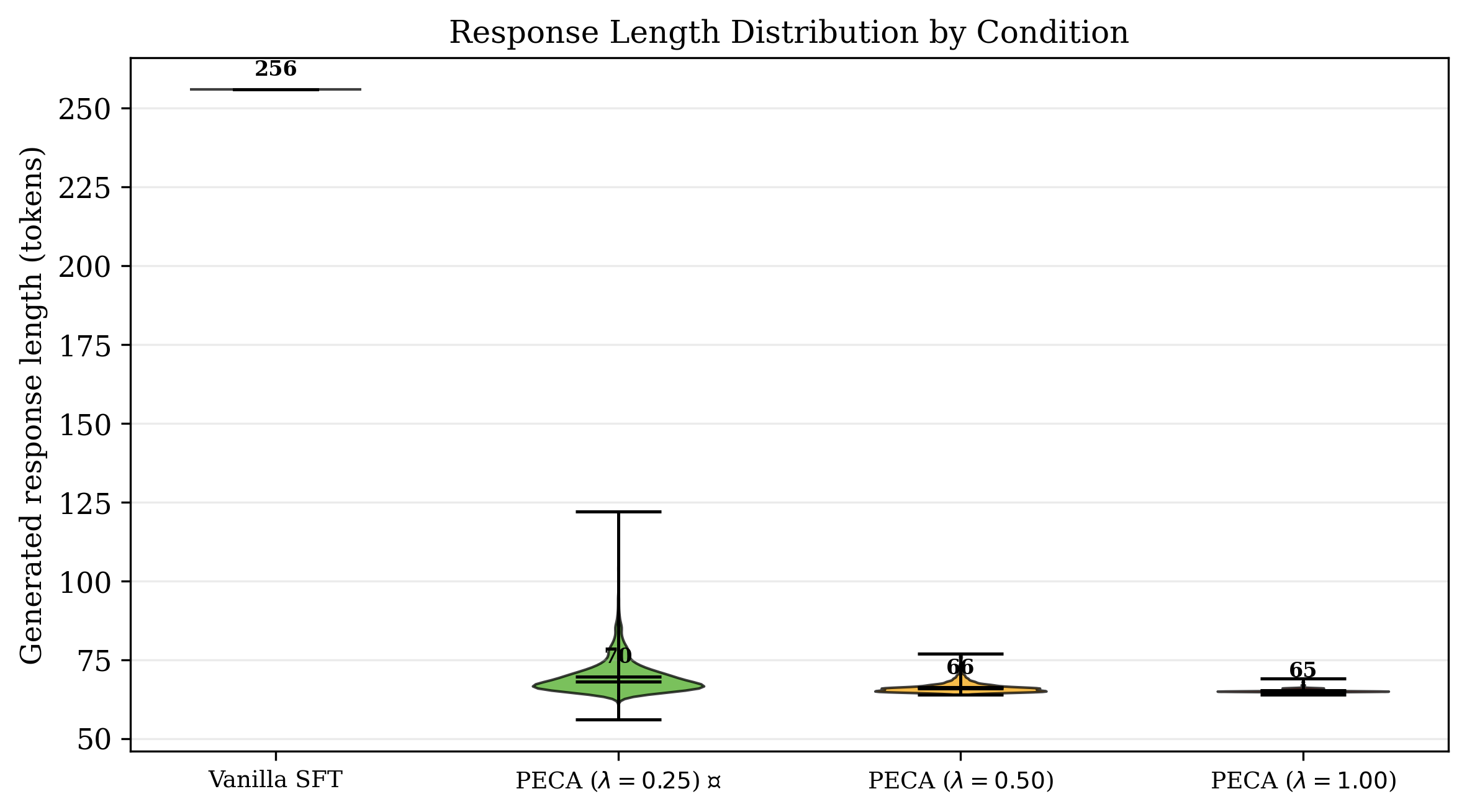}}{\fbox{\parbox[c][4cm][c]{0.8\linewidth}{\centering \textit{Figure file \texttt{length\_distribution.png} not found.}\\[4pt]\small Upload this file to the project to render the figure.}}}
\caption{Distributions of response length for vanilla SFT and the PECA
conditions; across every $\lambda$, PECA pulls the whole distribution toward
shorter outputs.}
\label{fig:length_dist}
\end{figure}

Fig.~\ref{fig:length_dist} shows the distribution of response lengths across the
test set. For vanilla SFT the lengths were spread around 256 tokens with a long
right tail. For every PECA setting the lengths were concentrated below 128 tokens.
What matters here is that the entire distribution slid left, not merely its long
tail. The model, in other words, learned when to stop rather than being clipped at
a fixed ceiling.

\subsection*{Training Dynamics}\label{subsec:dynamics}

\begin{figure}[ht]
\centering
\IfFileExists{loss_curves.png}{\includegraphics[width=0.72\linewidth]{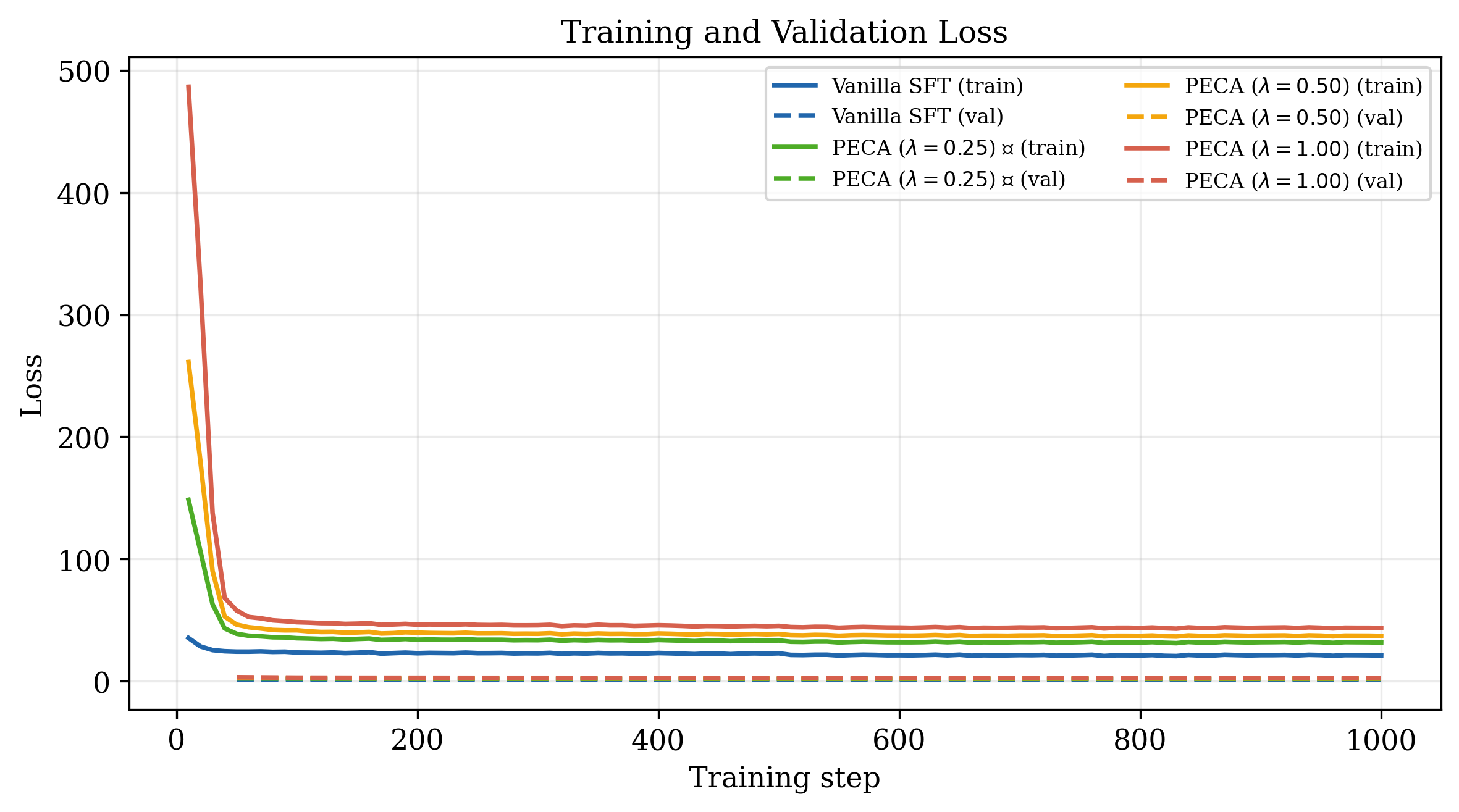}}{\fbox{\parbox[c][4cm][c]{0.8\linewidth}{\centering \textit{Figure file \texttt{loss\_curves.png} not found.}\\[4pt]\small Upload this file to the project to render the figure.}}}
\caption{Training and validation loss curves for all conditions over 2 epochs.
All conditions converge smoothly, and the PECA conciseness term does not
destabilize training.}
\label{fig:loss}
\end{figure}

The training and validation loss curves for every condition appear in
Figure~\ref{fig:loss}. Over the two epochs all four runs settle smoothly, and
their training and validation curves stay close to one another, which suggests
the QLoRA adapters generalise rather than memorise the 8,000 training examples.
Adding the conciseness term does nothing to upset this behaviour; the EOS-budget
objective slots into the ordinary supervised loop without any sign of
instability.

\subsection*{Computational Efficiency}\label{subsec:efficiency}

On top of the standard QLoRA pipeline, PECA introduces nothing but the
differentiable term in Eq.~\eqref{eq:conc}; there are no auxiliary models, no
reward functions, and no preference data to gather. Training each condition took
between 153 and 163 minutes on a single NVIDIA A10G or L4 GPU, with peak memory
use of 4.4 to 5.3~GB. That fits comfortably under the 16~GB available on
commonplace hardware, free-tier cloud GPUs included, so conciseness adaptation
turns out to be within reach for anyone, without specialised infrastructure.

\subsection*{Qualitative Analysis}\label{subsec:qualitative}

A single case shows why the shorter answer is worth more than the statistics alone
let on (Table~\ref{tab:qual}). A woman at 32 weeks presents with painful,
continuous vaginal bleeding. The vanilla model takes the differential one entry at
a time, and it is still weighing options when the 256-token ceiling arrives; the
answer never gets written. Nothing was wrong with the reasoning it produced, only
that the reasoning used up the whole budget and left no room for a conclusion. A
diagnosis a busy clinician never sees might as well not have been reached, because
it changes no decision.

At $\lambda=0.25$ the model walked the same path but finished in 68 tokens, a
73.4\% cut, and stated the diagnosis in one paragraph. It fixed on the feature that
decides the case, bleeding that hurts rather than bleeding that does not, and got
the answer right. Turn $\lambda$ up to 1.0 and there is no longer room for reasoning
at all, which is why accuracy all but vanishes there. In small, the case tracks the
aggregate: a light touch helped, and a heavy one cut into the reasoning itself.

\begin{table}[!ht]
\centering
\caption{Vanilla SFT and PECA ($\lambda=0.25$) side by side on one clinical
vignette, a 32-week pregnant patient with painful vaginal bleeding. Vanilla SFT
spends its full 256-token budget without ever naming a diagnosis; PECA reaches the
right one in 68 tokens, cutting length by 73.4\%.}
\label{tab:qual}
\begin{tabular}{p{0.17\linewidth}p{0.75\linewidth}}
\toprule
& \textbf{Output} \\
\midrule
Vanilla SFT
(256 tokens,
no diagnosis
reached) &
Alright, let's think this through. We've got a 32-week pregnant woman who's in
the ER with continuous bleeding from the vagina. That's definitely concerning.
Now, she's also saying the uterus is painful and tender. That's a big clue. So,
what could be causing this? First off, let's consider the options. A circumvallate
placenta is a type of placenta that's usually not a problem in the first trimester,
so it doesn't seem likely here. Rupture of membranes is more about the membranes
breaking, not necessarily causing bleeding like this. Placenta praevia would
typically present with painless bleeding \ldots\
[response truncated at generation limit; diagnosis never stated] \\
\addlinespace
PECA\\($\lambda$=0.25)
(68 tokens) &
Alright, so we have a 32-week pregnant woman in the ER with continuous vaginal
bleeding. The uterus is painful and tender, which is the key distinguishing sign.
Placenta praevia typically causes painless bleeding, so that is less likely here.
A painful, tender uterus with bleeding at this gestation points clearly toward
accidental haemorrhage (placental abruption). \\
\bottomrule
\end{tabular}
\end{table}

\subsection*{Comparison with Existing Approaches}\label{subsec:comparison}

PECA makes most sense set against the approaches that usually own this problem,
the preference-optimisation family and the reinforcement-learning family, plus the
ordinary PEFT baselines that never touch length at all. Table~\ref{tab:comparison}
lines them up on the criteria a low-resource clinical setting actually cares about:
the supervision each one demands, whether its length signal is differentiable
inside supervised fine-tuning, whether it costs anything at inference, and whether
it is parameter-efficient. This is a comparison of what each method \emph{is}, not
a leaderboard; because the entries were tested on different models, data, and
tasks, their published numbers do not line up with ours, and we make no pretence of
tabulating them side by side.

\begin{table}[!ht]
\centering
\caption{Characteristic comparison of PECA with representative length-control and
PEFT methods. ``Pref. data'' denotes a requirement for paired preference
annotations; ``Reward/teacher'' denotes a learned reward model, reference policy,
or external teacher model; ``Diff. SFT term'' indicates whether length is
controlled by a differentiable term inside ordinary supervised fine-tuning.}
\label{tab:comparison}
\setlength{\tabcolsep}{5pt}\renewcommand{\arraystretch}{1.15}\small
\resizebox{\textwidth}{!}{%
\begin{tabular}{lccccc}
\toprule
Method & Stage & Pref.\ data & Reward/teacher & Diff.\ term & Param.\ eff. \\
\midrule
Vanilla LoRA/QLoRA [2, 12] & SFT (PEFT)      & No  & No            & ---  & Yes \\
DPO, length-reg.\ [33]     & Preference opt. & Yes & Ref.\ policy  & No   & No  \\
SimPO [32]                 & Preference opt. & Yes & Ref.-free     & No   & No  \\
LMPO [19]                  & Preference opt. & Yes & Ref.-free     & No   & No  \\
EOS weighting [31]         & SFT             & No  & No            & Yes  & No\textsuperscript{*} \\
SmartThinker [37]          & RL (GRPO)       & No  & Length reward & No   & No  \\
Teacher distillation [11]  & SFT             & No  & Teacher LLM   & No   & Yes \\
\textbf{PECA (ours)}       & \textbf{SFT (PEFT)} & \textbf{No} & \textbf{No} & \textbf{Yes} & \textbf{Yes} \\
\bottomrule
\end{tabular}%
}

\vspace{2pt}
\raggedright\footnotesize\textsuperscript{*}Introduced for full fine-tuning of a
summarisation model rather than in a parameter-efficient setting.
\end{table}

Read down the final column of Table~\ref{tab:comparison}, PECA is the only method
that is at once preference-free, teacher-free, differentiable within supervised
fine-tuning, free of any inference-time overhead, and parameter-efficient. The
closest prior method conceptually is the end-of-sequence weighting of Belligoli
et al.\ [31], which shares the differentiable, supervised, preference-free
character of our term; the difference is that it was developed for full
fine-tuning of a summarisation model, whereas PECA gates the signal by a token
budget and situates it inside QLoRA so that it runs on commodity hardware for
clinical chain-of-thought models. Relative to the reinforcement-learning line,
SmartThinker [37] reports up to 52.5\% length reduction with accuracy gains on
mathematical-reasoning benchmarks, and our method achieves a larger relative
reduction (72.8\%) with a diagnostic-accuracy gain on a clinical benchmark; the
two figures are not directly comparable because the model families and tasks
differ, but the comparison does show that a simple supervised term can reach the
compression regime that RL methods target without their rollout and
reward-modelling machinery. Against the preference-optimisation methods
[19, 32, 33], the practical distinction is starker still: each of them presupposes
preference pairs or a reference policy that a small clinical project is unlikely
to have, whereas PECA needs only the supervised data already in hand. A fully
controlled empirical head-to-head, in which these baselines are re-trained on the
identical Qwen3-1.7B backbone and medical corpus, is beyond the scope of the
present study and is left as future work.

\section*{Discussion}\label{sec:conclusion}

We set out to deal with a practical obstacle in deploying specialised small
language models: after domain adaptation they tend to answer at a length that
does not suit practitioners who have to move quickly. Parameter-Efficient
Conciseness Adaptation (PECA) is our answer, a preference-free method built around
a differentiable, budget-gated signal on the end-of-sequence probability.
Penalising the raw token count will not work, since a discrete count carries no
gradient; PECA gets around this by reinforcing the stopping probability once
generation passes a budget, an objective that lives inside ordinary supervised
fine-tuning and costs almost nothing to add.

Working with a 1.7-billion-parameter model and 8,000 medical chain-of-thought
examples, we found a length-versus-accuracy trade-off orderly enough that a
practitioner can navigate it deliberately instead of by guesswork. Our recommended
setting, $\lambda=0.25$, trims mean length by 72.8\% and, to our own surprise,
lifts MedMCQA accuracy by 3.2 points over vanilla QLoRA fine-tuning. Larger
$\lambda$ keeps trimming the output, but now accuracy is the price, and at the
extreme the model falls apart. The cost side is worth stating plainly: no run took
more than three hours on one commodity GPU under 6~GB, and none of them called for
preference data, a second model, or a paid endpoint.

What we take to be the most telling implication is that a domain-adapted SLM's
verbosity is mostly a residue of its training distribution, not some necessary
companion of good reasoning. A gentle nudge toward stopping is enough to break the
habit, and doing so nudges multiple-choice accuracy upward rather than down, most
likely because the text being cut is hedging and repetition, words that pad a
response without adding anything a clinician could act on. The contribution of
PECA, in operational terms, is to expose a single interpretable parameter through
which a deployment can be placed at its preferred location on the brevity-accuracy
frontier.

\noindent\textbf{Limitations.} This work rests on a single random seed and a
single model family, and the absolute MedMCQA figures represent the capability
ceiling of a 1.7B model on a demanding benchmark rather than that of a fully
engineered clinical decision-support system. Validating outputs with clinicians,
extending PECA to larger model families and additional scales, and studying how
the token budget $B$ should be calibrated to the natural answer length of
different medical specialties are all important directions for future
investigation.

\section*{Methods}\label{sec:methodology}\addcontentsline{toc}{section}{Methods}

The Methods are organised as follows. We first describe the conciseness objective
and the training pipeline that distinguishes PECA from a standard supervised
baseline, then detail the medical chain-of-thought corpus and its preprocessing,
the backbone model and its 4-bit quantisation, the QLoRA adapter configuration,
the composite training objective together with the budget-gated end-of-sequence
term at its core, and finally the optimisation settings and reproducibility
details. Throughout, the baseline and PECA models are held identical in every
respect except the objective, so that any difference in behaviour can be
attributed to the conciseness term alone.

\subsection*{Parameter-Efficient Conciseness Adaptation (PECA)}\label{subsec:peca}

We present Parameter-Efficient Conciseness Adaptation (PECA), a framework that
addresses excessive response verbosity in domain-specific small language models,
a behaviour that impedes clinical use even when the underlying diagnostic
reasoning is sound [15, 18]. In contrast to approaches that rely on
resource-intensive preference optimisation, such as Direct Preference
Optimization [14, 33], SimPO [32], and LMPO [19], or on distillation from
proprietary large language models [11], PECA achieves clinically appropriate
brevity through a single, fully differentiable term added to standard supervised
fine-tuning (SFT) under quantized low-rank adaptation (QLoRA) [2, 12].

\begin{figure}[!ht]
\centering
\IfFileExists{PECA_figure1.png}{\includegraphics[width=0.95\linewidth]{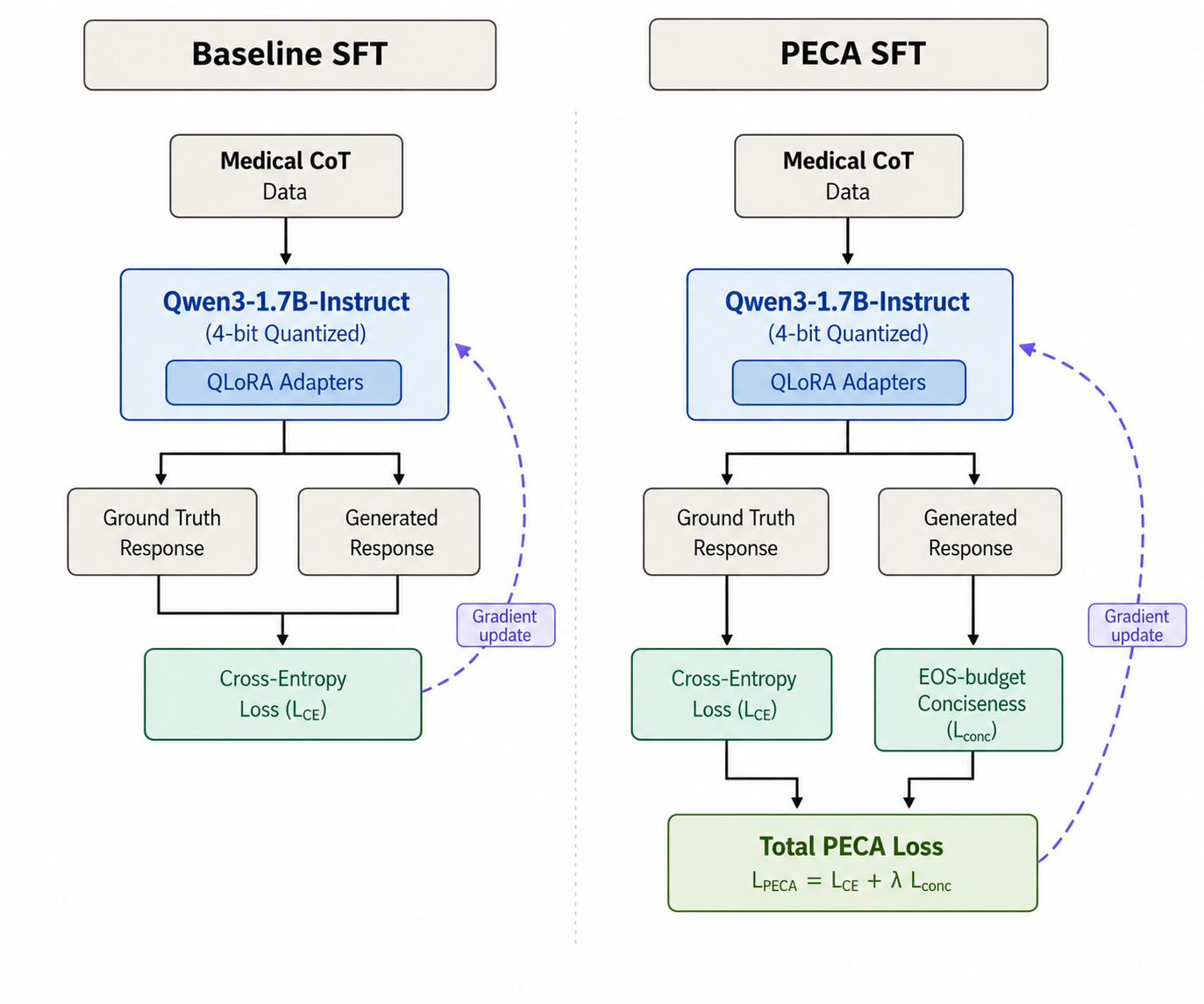}}{\fbox{\parbox[c][4cm][c]{0.8\linewidth}{\centering \textit{Figure file \texttt{PECA\_figure1.png} not found.}\\[4pt]\small Upload this file to the project to render the figure.}}}
\caption{Overview of the Baseline SFT and the proposed PECA SFT training setup.
Both share a 4-bit quantized Qwen3-1.7B-Instruct backbone with QLoRA adapters
and differ only in the training objective: the baseline uses cross-entropy alone,
while PECA adds the differentiable EOS-budget conciseness term
$\mathcal{L}_{\mathrm{conc}}$.}
\label{fig:peca_architecture}
\end{figure}

What defines PECA is how little it asks for. There is no preference data to
collect, no helper model at inference, no paid interface, so the whole thing
reproduces on commodity hardware. Figure~\ref{fig:peca_architecture} shows the
layout. Baseline and PECA travel the same pipeline over a 4-bit quantized
Qwen3-1.7B-Instruct backbone with QLoRA adapters, and the objective is the single
point of difference between them. Holding everything else constant is what lets us
pin any shift in generation behaviour on the conciseness term, and what makes the
comparison a controlled one.

Both start from the same 4-bit quantized Qwen3-1.7B-Instruct backbone and train
the same QLoRA adapters. Only the objective sets them apart: the baseline
optimises cross-entropy $\mathcal{L}_{\mathrm{CE}}$ alone, while PECA adds a
differentiable conciseness term $\mathcal{L}_{\mathrm{conc}}$ that acts on the
response tokens only.

\subsection*{Dataset and Preprocessing}\label{subsec:data}

The data are the \texttt{medical-o1-reasoning-SFT} corpus [34, 35], an open
medical chain-of-thought set on Hugging Face under Apache-2.0. Its problems come
from MedQA-USMLE and MedMCQA, each one verifiable; a capable teacher model writes
a long reasoning trace for the problem, and the trace survives only if a medical
verifier accepts it. Three fields make up each instance, a clinical
\emph{Question}, a lengthy \emph{Complex\_CoT} trace, and a short
\emph{Response}, and the English split leaves us 19,704 usable examples.

\begin{table}[h!]
\centering
\caption{Characteristics of the \texttt{medical-o1-reasoning-SFT} (English)
dataset.}
\label{tab:medical-cot}
\begin{tabular}{@{}lr@{}}
\hline
\textbf{Attribute}                       & \textbf{Value}              \\
\hline
Total usable instances                   & 19,704                      \\
Average question length (words)          & 46.11                       \\
Average CoT reasoning length (words)     & 312.45                      \\
Average response length (words)          & 93.98                       \\
Reasoning generation                     & Teacher LLM, verifier-validated \\
Source problems                          & MedQA-USMLE, MedMCQA        \\
License                                  & Apache-2.0                  \\
\hline
\end{tabular}
\end{table}

We sample 8,000 examples at random from that English split to train on, hold out
1,000 for validation and another 1,000 for the final evaluation, and leave the
remainder aside. The validation set is where $\lambda$, the token budget $B$, and
the rest of the hyperparameters get settled; the test set we open only for the
numbers we report.

\begin{table}[h!]
\centering
\caption{Dataset splits used in this study.}
\label{tab:splits}
\begin{tabular}{lc}
\toprule
Split      & Instances \\
\midrule
Training   & 8,000     \\
Validation & 1,000     \\
Test       & 1,000     \\
\bottomrule
\end{tabular}
\end{table}

Each example is wrapped in a fixed prompt that holds the reasoning and the final
answer in separate slots:
\begin{quote}
\textbf{Instruction}: You are a medical expert. Think through the case step by
step, then give a concise final answer.\\
\textbf{Question}: [clinical presentation]\\
\textbf{Reasoning and Answer}: [diagnostic reasoning followed by final answer]
\end{quote}

Tokenisation goes through the official Qwen3 tokenizer, with sequences capped at
768 tokens. Table~\ref{tab:token-distribution} shows that the three splits share
very similar token-length distributions, which reassures us that the random
sampling left the partitions balanced.

\begin{table}[h!]
\centering
\caption{Token length distribution across dataset splits after preprocessing.}
\label{tab:token-distribution}
\begin{tabular}{lccc}
\toprule
Split      & Mean   & Median & 95th Percentile \\
\midrule
Training   & 523.7  & 507.5  & 686.1           \\
Validation & 525.9  & 509.0  & 701.1           \\
Test       & 525.2  & 510.5  & 691.1           \\
\bottomrule
\end{tabular}
\end{table}

\subsection*{Model Architecture and Quantization}\label{subsec:arch}

The base model is Qwen3-1.7B-Instruct [21], a 1.7-billion-parameter decoder-only
transformer that offers strong instruction-following and multi-step reasoning
among open-weight models in the sub-2B class, while remaining small enough to
fine-tune on a single consumer GPU. The model is quantized to 4-bit precision
using the NormalFloat4 (NF4) datatype with double quantization [12], reducing
the memory footprint to approximately 1.1~GB and enabling QLoRA fine-tuning well
within a 16~GB GPU budget.

\subsection*{QLoRA Configuration}\label{subsec:qlora}

Low-rank adaptation is implemented with the QLoRA framework [12] using the
configuration in Table~\ref{tab:lora-config}, applied uniformly to the baseline
and PECA models so that any behavioural difference is attributable to the
objective alone.

\begin{table}[h!]
\centering
\caption{QLoRA adapter configuration.}
\label{tab:lora-config}
\begin{tabular}{lc}
\toprule
Hyperparameter       & Value \\
\midrule
Rank ($r$)           & 16    \\
Scaling factor ($\alpha$) & 32 \\
Dropout              & 0.05  \\
Target modules       & q\_proj, k\_proj, v\_proj, o\_proj, \\
                     & gate\_proj, up\_proj, down\_proj \\
Trainable parameters & 17,432,576 \\
Percentage of total  & 1.69\% of 1,033,364,480 \\
\bottomrule
\end{tabular}
\end{table}

\subsection*{Training Objectives}\label{subsec:objective}

The baseline model is trained with standard next-token prediction using
cross-entropy loss over the supervised tokens:
\begin{equation}
    \mathcal{L}_{\mathrm{CE}} = -\sum_{t=1}^{T} \log P_{\theta}(y_t \mid y_{<t}, x).
    \label{eq:ce}
\end{equation}

\paragraph{Why a naive length penalty does not work.}
A tempting approach would be to add a term proportional to the realized output
length, such as $\max(0,|y|-L_{\max})/L_{\max}$. However, the number of
generated tokens $|y|$ is a discrete quantity. Under teacher-forced supervised
fine-tuning this count is piecewise-constant with respect to the model parameters
$\theta$, so its gradient is zero almost everywhere; such a term cannot drive
learning and would leave the model's generation length essentially unchanged. We
therefore realize conciseness through a differentiable surrogate that operates
on the EOS probability rather than on the token count.

\paragraph{Budget-gated EOS conciseness term.}
Let $\mathcal{R}$ denote the set of token positions in the supervised response
span (the tokens following the final answer delimiter), and let
$P_{\theta}(\mathrm{EOS}\mid y_{<t}, x)$ be the model's probability of emitting
EOS at position $t$. Given a token budget $B$, define the conciseness term as
the mean negative log-probability of stopping at every response position beyond
the budget:
\begin{equation}
    \mathcal{L}_{\mathrm{conc}}
    = -\frac{1}{|\mathcal{R}_{>B}|}\sum_{t \in \mathcal{R},\, t > B}
      \log P_{\theta}\!\left(\mathrm{EOS}\mid y_{<t}, x\right),
    \qquad \mathcal{R}_{>B} = \{\, t \in \mathcal{R} : t > B \,\}.
    \label{eq:conc}
\end{equation}
Minimizing Eq.~\eqref{eq:conc} increases the EOS probability at every response
position past the budget $B$, encouraging the model to terminate once a
clinically sufficient answer has been produced. Because it rides on the softmax
EOS probability, the term varies smoothly, differentiates cleanly, and adds next
to nothing to the forward and backward passes already being computed. Adding it to
the cross-entropy loss yields the composite PECA objective
\begin{equation}
    \mathcal{L}_{\mathrm{PECA}} = \mathcal{L}_{\mathrm{CE}} + \lambda\,\mathcal{L}_{\mathrm{conc}},
    \label{eq:peca}
\end{equation}
Here $\lambda \geq 0$ sets how hard the conciseness pressure pushes and $B$ marks
the target length. We run $\lambda \in \{0.25, 0.50, 1.00\}$ alongside the vanilla
baseline ($\lambda=0$), keeping $B=64$ fixed across the sweep.

\subsection*{Training and Optimization Configuration}\label{subsec:train}

We train for 2 epochs with the Paged AdamW 8-bit optimizer [22], a cosine
learning-rate schedule, and a 3\% warmup. The full configuration is given in
Table~\ref{tab:training-config}. Each condition is trained with seed 42; the
effect of the conciseness term is assessed across four $\lambda$ values rather
than across seeds, as the lambda sweep directly characterises the Pareto
frontier that is the primary contribution.

\begin{table}[h!]
\centering
\caption{Training hyperparameters.}
\label{tab:training-config}
\begin{tabular}{lc}
\toprule
Hyperparameter             & Value \\
\midrule
Optimizer                  & Paged AdamW 8-bit \\
Learning rate              & $2 \times 10^{-4}$ \\
Weight decay               & 0.01 \\
Effective batch size       & 16 \\
Gradient accumulation steps & 16 \\
Max sequence length        & 768 \\
Epochs                     & 2 \\
Precision                  & bfloat16 \\
Gradient clipping          & 0.3 \\
LoRA rank / alpha          & 16 / 32 \\
Token budget $B$           & 64 \\
Random seed                & 42 \\
\bottomrule
\end{tabular}
\end{table}

\subsection*{Implementation and Reproducibility}\label{subsec:repro}

All experiments were run on NVIDIA A10G and L4 GPUs using PyTorch, Hugging Face
Transformers, PEFT, and BitsAndBytes. Peak GPU memory across all conditions was
4.4 to 5.3~GB, well within a 16~GB budget, confirming that PECA is accessible
on widely available hardware. Each condition completed in approximately 153 to
163 minutes. The full source code, processed dataset splits,
training logs, and trained QLoRA adapters will be made publicly available upon
publication at
\url{https://github.com/tanzeelabbas/PECA-Medical} and on Hugging Face at
\url{https://huggingface.co/tanzeelabbas114}.

\section*{Data availability}\label{sec:data-avail}
The medical-o1-reasoning-SFT dataset analysed in this study is publicly available
on Hugging Face at
\url{https://huggingface.co/datasets/FreedomIntelligence/medical-o1-reasoning-SFT}.
The processed dataset splits and trained adapters generated during this study will
be made publicly available upon publication at
\url{https://github.com/tanzeelabbas/PECA-Medical} and on Hugging Face at
\url{https://huggingface.co/tanzeelabbas114}. During peer review they can be made
available to the editor and reviewers on request.

\section*{Funding Declaration}\label{sec:funding}
This research work is partially funded by the School of Health and Behavioral
Sciences, Bryant University, United States.

\section*{Use of large language models}
A large language model based assistant was used for help with code
implementation and with language editing of portions of the manuscript. The
experimental design, analysis, and all scientific conclusions are the authors'
own, and every claim, number, and result was checked and verified by the authors,
who take full responsibility for the accuracy and integrity of the work.


\begin{thebibliography}{99}

\bibitem{ref1}
Ding, N., Qin, Y., Yang, G., Wei, F., Yang, Z., Su, Y., Hu, S., Chen, Y.,
Dai, D., Huang, L., Wang, H., Xu, J., Han, J., Su, H., Zhou, H., Zhang, T.,
Liu, Z., Sun, M.: Parameter-efficient fine-tuning of large-scale pre-trained
language models. Nat. Mach. Intell. 5(3), 220--235 (2023)

\bibitem{ref2}
Hu, E.J., Shen, Y., Wallis, P., Allen-Zhu, Z., Li, Y., Wang, S., Wang, L.,
Chen, W.: LoRA: Low-rank adaptation of large language models.
arXiv preprint arXiv:2106.09685 (2021)

\bibitem{ref3}
He, Y., Zhu, X., Li, D., Wang, H.: Enhancing large language models for
specialized domains: A two-stage framework with parameter-sensitive LoRA
fine-tuning and chain-of-thought RAG. Electronics 14(10), 1961 (2025)

\bibitem{ref4}
Lu, W., Luu, R.K., Buehler, M.J.: Fine-tuning large language models for domain
adaptation: Exploration of training strategies, scaling, model merging and
synergistic capabilities. npj Comput. Mater. \textbf{11}, 84 (2025)

\bibitem{ref5}
Liang, X., Khaw, Y.-M.J., Liew, S.-Y., Tan, T.-P., Qin, D.: Toward low-resource
languages machine translation: A language-specific fine-tuning with LoRA for
specialized large language models. IEEE Access 13, 46616--46627 (2025)

\bibitem{ref6}
Anisuzzaman, D.M., Malins, J.G., Friedman, P.A., Attia, Z.I.: Fine-tuning large
language models for specialized use cases. Mayo Clin. Proc.: Digit. Health 3(1),
100184 (2025)

\bibitem{ref7}
Fatemi, S., Hu, Y., Mousavi, M.: A comparative analysis of instruction
fine-tuning large language models for financial text classification. arXiv
preprint arXiv:2411.02476 (2024)

\bibitem{ref8}
Olariu, I., Lothritz, C., Klein, J., Bissyand\'e, T., Guo, S., Haddadan, S.:
Evaluating parameter-efficient finetuning approaches for pre-trained models on
the financial domain. In: Findings of EMNLP 2023, pp. 15482--15491 (2023)

\bibitem{ref9}
Yan, Y., Yu, H., Wang, D., Ye, J., Liu, F., Xu, W.: STP: Special token prompt
for parameter-efficient tuning of pre-trained language models.
Expert Syst. Appl. 265, 127665 (2025)

\bibitem{ref10}
Prottasha, N.J., Mahmud, A., Sobuj, M.S.I., Bhat, P., Kowsher, M., Yousefi, N.,
Garibay, O.O.: Parameter-efficient fine-tuning of large language models using
semantic knowledge tuning. Sci. Rep. \textbf{14}, 30667 (2024)

\bibitem{ref11}
Liu, Q., Wu, X., Zhao, X., Zhu, Y., Xu, D., Tian, F., Zheng, Y.: When MOE
meets LLMs: Parameter-efficient fine-tuning for multi-task medical
applications. In: Proceedings of the 47th International ACM SIGIR Conference
on Research and Development in Information Retrieval, pp. 1104--1114 (2024)

\bibitem{ref12}
Dettmers, T., Pagnoni, A., Holtzman, A., Zettlemoyer, L.: QLoRA: Efficient
finetuning of quantized LLMs. arXiv preprint arXiv:2305.14314 (2023)

\bibitem{ref13}
Houlsby, N., Giurgiu, A., Jastrzebski, S., Morrone, B., De Laroussilhe, Q.,
Gesmundo, A., Attariyan, M., Gelly, S.: Parameter-efficient transfer learning
for NLP. In: ICML, vol. 97, pp. 2790--2799 (2019)

\bibitem{ref14}
Rafailov, R., Sharma, A., Mitchell, E., Ermon, S., Manning, C.D., Finn, C.:
Direct preference optimization: Your language model is secretly a reward model.
In: NeurIPS (2023)

\bibitem{ref15}
He, H., Cai, J., Zhang, J., Tao, D., Zhuang, B.: Sensitivity-aware visual
parameter-efficient fine-tuning. In: ICCV, pp. 11825--11835 (2023)

\bibitem{ref16}
Nguyen, V., Karimi, S., Rybinski, M., Xing, Z.: MedRedQA for medical consumer
question answering: Dataset, tasks, and neural baselines. In: Proceedings of
IJCNLP-AACL 2023 (Volume 1: Long Papers), pp. 629--648 (2023)

\bibitem{ref17}
Phogat, K.S., Puranam, S.A., Dasaratha, S., Harsha, C., Ramakrishna, S.:
Fine-tuning smaller language models for question answering over financial
documents. arXiv preprint arXiv:2408.12337 (2024)

\bibitem{ref18}
Kim, H., Hwang, H., Lee, J., Park, S., Kim, D., Lee, T., Yoon, C., Sohn, J.,
Park, J., Kang, J.: Small language models learn enhanced reasoning skills from
medical textbooks. npj Digit. Med. \textbf{8}, 240 (2025)

\bibitem{ref19}
Li, G., Xia, T., Chang, Y., Wu, Y.: Length-controlled margin-based preference
optimization (LMPO) without reference model. arXiv preprint arXiv:2502.14643
(2025)

\bibitem{ref20}
Zhang, T., Kishore, V., Wu, F., Weinberger, K.Q., Artzi, Y.: BERTScore:
Evaluating text generation with BERT. In: International Conference on Learning
Representations (ICLR) (2020)

\bibitem{ref21}
Qwen Team: Qwen3 technical report. arXiv preprint arXiv:2505.09388 (2025)

\bibitem{ref22}
Loshchilov, I., Hutter, F.: Decoupled weight decay regularization (AdamW).
In: ICLR (2019)

\bibitem{ref23}
Agarwal, A., Ramesh, S.K., Sengupta, A., Chakraborty, T.: Step-by-step
unmasking for parameter-efficient fine-tuning of large language models.
arXiv preprint arXiv:2408.14470 (2024)

\bibitem{ref24}
Li, Y., Yan, Y., Tong, Z., Wang, Y., Yang, Y., Bai, M., Pu, D., Xie, J.,
Liu, C., Li, B., Liu, M., Shu, K.: Efficient fine-tuning of small-parameter
large language models for biomedical bilingual multi-task applications.
Appl. Soft Comput. \textbf{175}, 113084 (2025)

\bibitem{ref25}
Deng, Y., Zhou, G., Zhao, Q.: RaLo: Rank-aware low-rank adaptation for
pre-trained foundation models. Neural Networks, 108423 (2025)

\bibitem{ref26}
Kumar, S., Ranjan, V., Chakrabarti, A., Das, T.K., Singh, A.: Efficient
biomedical text summarization with quantized LLaMA 2: Enhancing memory usage
and inference on low-powered devices. Expert Systems \textbf{42}(2), e13760
(2025)

\bibitem{ref27}
Dai, X., Zhou, Y., Qiu, X., Tang, H., Tong, T.: Parameter-efficient fine-tuning
for single image snow removal. Expert Syst. Appl. \textbf{265}, 125993 (2025)

\bibitem{ref28}
Tian, C., Shi, Z., Guo, Z., Li, L., Xu, C.-Z.: HydraLoRA: An asymmetric LoRA
architecture for efficient fine-tuning. In: Advances in Neural Information
Processing Systems \textbf{37}, pp. 9565--9584 (2024)

\bibitem{ref29}
Liu, H., Chen, Z., Li, P., Liu, Y.-Z., Liu, X., Xu, R.X., Sun, M.:
Resource-efficient instruction tuning of large language models for biomedical
named entity recognition. J. Biomed. Inform., 104896 (2025)

\bibitem{ref30}
Sun, G., Khalid, U., Mendieta, M., Wang, P., Chen, C.: Exploring
parameter-efficient fine-tuning to enable foundation models in federated
learning. In: 2024 IEEE International Conference on Big Data (BigData).
IEEE (2024)

\bibitem{ref31}
Belligoli, Z., Stergiadis, E., Fainman, E., Gusev, I.: Controlling
summarization length through EOS token weighting. arXiv preprint
arXiv:2506.05017 (2025)

\bibitem{ref32}
Meng, Y., Xia, M., Chen, D.: SimPO: Simple preference optimization with a
reference-free reward. In: NeurIPS (2024). arXiv:2405.14734

\bibitem{ref33}
Park, R., Rafailov, R., Ermon, S., Finn, C.: Disentangling length from quality
in direct preference optimization. In: Findings of the ACL (2024).
arXiv:2403.19159

\bibitem{ref34}
Chen, J., Ouyang, R., Gao, A., Chen, S., Chen, G.H., Wang, X., Zhang, R.,
Cai, Z., Ji, K., Yu, G., Wan, X., Wang, B.: HuatuoGPT-o1: Towards medical
complex reasoning with LLMs. arXiv preprint arXiv:2412.18925 (2024)

\bibitem{ref35}
FreedomIntelligence: medical-o1-reasoning-SFT dataset. Hugging Face (2025).
\url{https://huggingface.co/datasets/FreedomIntelligence/medical-o1-reasoning-SFT}

\bibitem{ref36}
Liu, G., Tang, X., He, J., Li, P., Chen, Z., Zhong, S.: PeFoMed:
Parameter-efficient fine-tuning of multimodal large language models for medical
CXR. Sci. Rep. \textbf{16}, 18729 (2026). \url{https://doi.org/10.1038/s41598-026-47871-2}

\bibitem{ref37}
Hu, C., Hu, Q., Xu, Y., Chen, J., Wang, R., Liu, S., Li, J., Wu, F., Chen, G.:
SmartThinker: Progressive chain-of-thought length calibration for efficient large
language model reasoning. arXiv preprint arXiv:2603.08000 (2026)


\bibitem{ref38}
Chen, X. et al. Grounding large language models in clinical diagnostics.
Nat. Commun. \textbf{17}, 4401 (2026)

\end{thebibliography}
\end{document}